\documentclass[runningheads]{llncs}
\usepackage[T1]{fontenc}
\usepackage{comment}
\usepackage{amsmath}
\usepackage{graphicx}
\usepackage{multirow}
\usepackage{paralist}           
\usepackage{pgfplots}
\usepackage{pgf-pie} 
\usepackage{booktabs}
\pgfplotsset{compat=1.18}

\makeatletter
\newcommand{\printfnsymbol}[1]{%
  \textsuperscript{\@fnsymbol{#1}}%
}
\makeatother
\begin{document}
%
\title{A Framework for Adapting Human-Robot Interaction to Diverse User Groups}
\titlerunning{A Framework for Adaptive Human-Robot Interaction}
%
\author{Theresa {Pekarek Rosin}\inst{1}\thanks{These authors contributed equally to this work and share first authorship.} \and
Vanessa Hassouna\inst{2}\printfnsymbol{1} \and
Xiaowen Sun\inst{1}\printfnsymbol{1} \and Luca Krohm\inst{2} \and Henri-Leon Kordt\inst{1} \and Michael Beetz\inst{2} \and Stefan Wermter\inst{1}}
\authorrunning{T. Pekarek Rosin et al.}
%
\institute{Knowledge Technology, Department of Informatics, University of Hamburg,
Vogt-Koelln-Str. 30, 22527 Hamburg, Germany\\
\email{\{theresa.pekarek-rosin, xiaowen.sun, henri-leon.kordt, stefan.wermter\}@uni-hamburg.de}\\
\url{www.knowledge-technology.info} \and
Institute of Artificial Intelligence, University Bremen, Am Fallturm 1, 
28359 Bremen, Germany\\
\email{\{hassouna, beetz, luc\_kro\}@uni-bremen.de} \\
 \url{www.ai.uni-bremen.de}}
\maketitle              
\begin{abstract}
To facilitate natural and intuitive interactions with diverse user groups in real-world settings, social robots must be capable of addressing the varying requirements and expectations of these groups while adapting their behavior based on user feedback. While previous research often focuses on specific demographics, we present a novel framework for adaptive Human-Robot Interaction (HRI) that tailors interactions to different user groups and enables individual users to modulate interactions through both minor and major interruptions.
Our primary contributions include the development of an adaptive, ROS-based HRI framework with an open-source code base. This framework supports natural interactions through advanced speech recognition and voice activity detection, and leverages a large language model (LLM) as a dialogue bridge. We validate the efficiency of our framework through module tests and system trials, demonstrating its high accuracy in age recognition and its robustness to repeated user inputs and plan changes.

\keywords{Social Robotics  \and Symbolic Planning  \and Age Recognition \and Large Language Models\and Human-Robot Interaction 
}
\end{abstract}

\section{Introduction}
The field of Human-Robot Interaction (HRI) has often focused on examining specific demographic groups, such as the elderly and children, separately, due to their unique interaction dynamics~\cite{Asgharian2022,mordoch2013social,Raptopulou2021}. 
However, in real-life environments, a diverse range of people often live and work together, and social robots need to adapt to different demands and expectations~\cite{Joshi2019,Mutlu2008}.
Given that not every person interacting with a social robot is an experienced user, the interaction design must prioritize usability principles. These include efficiency of use, minimization of cognitive load, consistency, feedback, error prevention, and ethical considerations such as information privacy and maintaining user control~\cite{Fronemann2021}.

Voice interaction is particularly effective at lowering cognitive load. 
It offers intuitive use and efficiency, does not require expert knowledge from the user~\cite{Robinson2022,Tellex2020}, and is generally better received than other communication modes due to the familiar nature of vocal feedback from robots~\cite{GUTMAN2023}. Speech also provides a range of paralinguistic cues, such as pitch, articulation, timing, and voice quality, which contain additional information about the user currently interacting with the system \cite{Ashok2022}. This has previously been used to adapt the robot's behavior based on the detected emotion of the user~\cite{Ashok2022}, but more general cues, such as age, can also be used to personalize the robot's behavior for different user groups.

Vocal feedback from the robot can also increase interaction transparency and reduce the black-box effect. Ideally, the robot should provide explanations of underlying decisions and processes tailored to the user~\cite{Nair2023}. 
To obtain the full benefit of a transparent robot, we believe that explanations should always be paired with repair mechanisms. Previous work on action or trust repair has focused on post-hoc repair mechanisms, usually for scenarios where social interaction was not the main focus~\cite{vanWaveren2022}. However, waiting until the end of an interaction to fix the robot's behavior could leave the user feeling stuck in a faulty interaction, which could lead to an increased perceived loss of control. For social scenarios, automatic repair mechanisms have been examined, which increase user satisfaction when combined with sufficient explanations but also circumvent the user's autonomy~\cite{Lee2024}.
Therefore, we argue that it is vital to allow users to interrupt the robot at any time to avoid situations the user is uncomfortable with, to adjust the actions to their preferences, or to perform trust, language, and action repair~\cite{Lee2024,Zhang2023}. 

Achieving this level of reactivity and flexibility is challenging for logic-based systems and usually requires expert knowledge to design the system for active inference~\cite{Pezzato2023}. However, large language models (LLMs), such as the various GPT models~\cite{radford2019language} or the LLaMa model series~\cite{touvron2023llama}, offer a new way to build flexible HRI scenarios~\cite{sun2024details,wang2024lami}. We believe that LLMs can bridge the gap in the communication between robot and user by performing the necessary integration of natural language queries and generating appropriate responses~\cite{Tellex2020}, without specifying and preparing for all contingencies.

In this paper, we introduce a framework for handling speech and natural language-based user interruptions in HRI within a simulated kitchen environment. We categorize interruptions into minor (plan changes) and major (complete stopping of the robot).
We argue that incorporating user-specific traits as dialogue and interaction-modulating context variables provides an intuitive approach to HRI for diverse user groups based on previous research for specific demographics, e.g. senior citizens~\cite{Asgharian2022,GUTMAN2023}. We utilize the user's age to adapt the interaction: For older adults, the interaction is simplified, and the behavior of the robot is more predictable, for example, through frequent vocalization of intent, while the interaction focuses on efficiency for younger people.

Our main contributions include an adaptive ROS-based framework for HRI with an open-source code base\footnote{\url{https://github.com/TPekarekRosin/UHH\_UB\_AgeAwareHRI}} that extends the PyCRAM language~\cite{kazhoyan17designators} with an Interrupt Client and recovery behavior. This framework enables natural user interactions through speech recognition and voice activity detection and implements an LLM as a dialogue bridge between the user and the robot.

We evaluate each module (speech/age recognition, dialogue bridge, robot planner) independently, and our entire architecture through system trials, where we demonstrate the effectiveness of our approach for two scenarios: fetching and replacing objects with minor interruptions, and setting the table and stopping the system with major interruptions.


\section{Related Work}
\begin{figure}[t]
\centering
\includegraphics[width=0.49\textwidth]{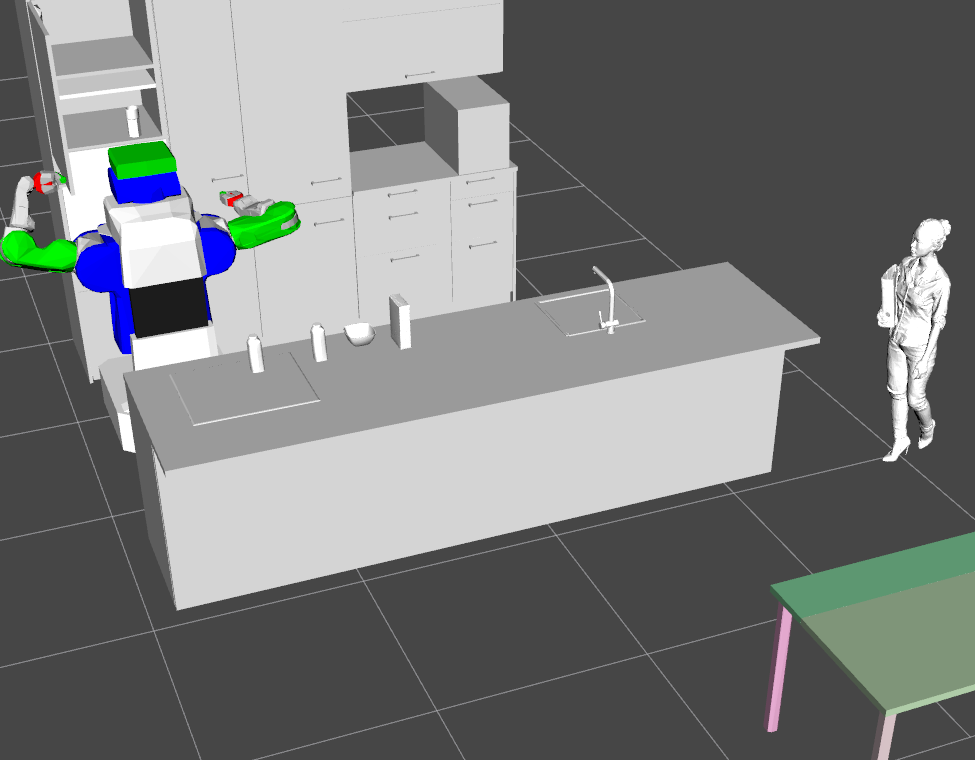} 
\includegraphics[width=0.49\textwidth]{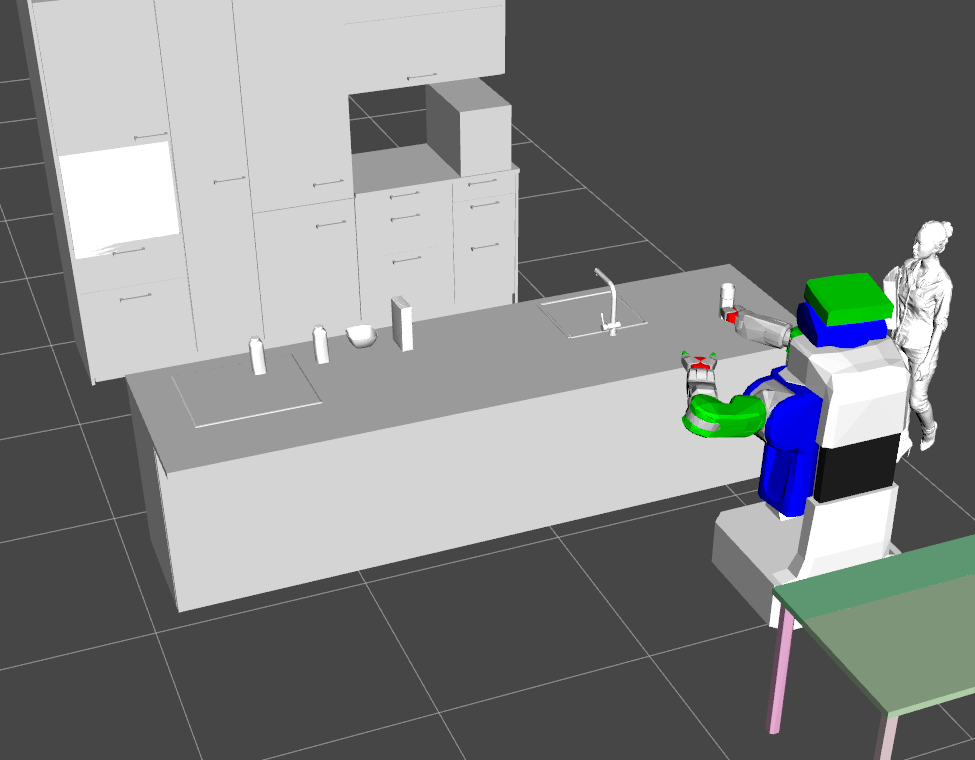}
\caption{The simulation environment with the kitchen scenario. Left: The robot moves around and interacts with the environment to search for the object. Right: It then places the object in front of the user on either the table or the counter.} 
\label{fig:simulation_environment}
\end{figure}
Interrupting and redirecting a robot's actions is crucial for handling failures in HRI. These failures must be communicated, perceived, and efficiently resolved by the robotic agent~\cite{Honig2018}. For instance, the ``Robot Household Marathon Experiment''~\cite{Kazhoyan2021RobotHousehold} highlights the importance of robots recovering from failures in real-world settings. Lee et al.~\cite{Lee2024} show that combining repair mechanisms with explanations enhances user trust and satisfaction. Feedback is vital to the recovery process and can be provided either through speech or system-specific feedback loops~\cite{Lee2024,vanWaveren2022}.

Even before large language models (LLMs), user feedback has been used for reactive action planning in robotic systems~\cite{Pezzato2023}. However, integrating LLMs into HRI has simplified handling requirements, such as natural language understanding, reasoning, and natural language generation~\cite{sun2024details,wang2024lami,Zhao2023}.
Bärmann et al.~\cite{bärmann2024} implement an HRI scenario in which human instructions or observations feed into an LLM (GPT-3.5/GPT-4), which learns incrementally from the feedback of a second LLM that adapts the prompt based on the previous (faulty) interaction. 
Similarly, Ye et al.~\cite{Ye2023} use ChatGPT to control a robot arm with natural language instructions from humans, finding that the LLM's understanding of human language nuances facilitates natural interactions.

These examples demonstrate the value of LLMs in HRI. However, even in multi-user scenarios~\cite{Lee2024}, there is an assumption that all users have the same needs and that all tasks require uniform levels of autonomy from the robot. Feedback is typically integrated post-hoc, despite findings from Gutman et al.~\cite{GUTMAN2023} suggesting that highly autonomous robots should accept user feedback during the interaction to mitigate the perceived loss of control. Additionally, since transient information like emotion~\cite{Ashok2022} is frequently used to modify robot behavior, constant factors such as age should be used to tailor interactions for specific user groups.

\section{Approach}

\begin{figure}[t]
\centering
\includegraphics[width=1\textwidth]{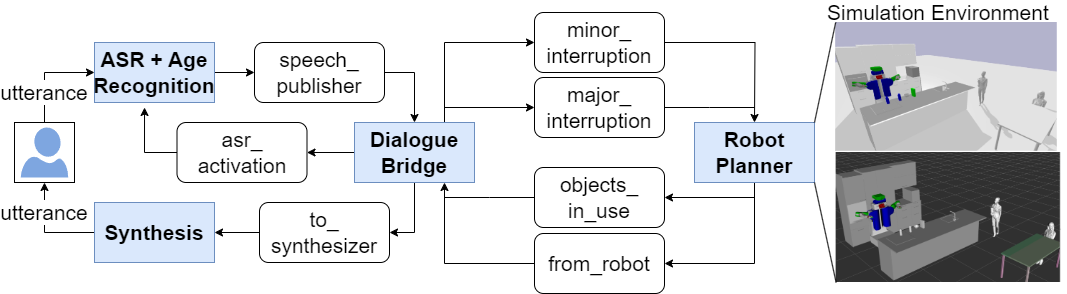}
\caption{Our architecture and the ROS communication processes. The user interacts with the system using natural language and receives vocal feedback. The user's speech is processed by an age and speech recognition model which transcribes the speech and detects the age group. This information is sent to the dialogue bridge, where commands and parameters are extracted and forwarded to the robotic agent, which executes the actions. The user can interrupt the robot at any time.} \label{fig:structure}
\end{figure}
Our interaction setup is a kitchen environment (Figure~\ref{fig:simulation_environment}), which we implement in the simulation environment BulletWorld\footnote{\url{https://www.cram-system.org/doc/pycram/bulletworld}}.
Users can request specific items (`milk', `bowl', etc.) or ask the robot to prepare breakfast, which triggers a sequence of actions to set the table. Our framework utilizes age recognition to initially configure the interaction, modifying the robot's actions and feedback based on the user's age. For older users, the robot more frequently vocalizes its next steps and movements, addressing the common difficulty they face in predicting the robot's actions~\cite{Fronemann2021}. These age-related adjustments form the basis for personalized interaction within the HRI framework, and user feedback is then used to adapt the robot's behavior to the individual user during the interaction.
We utilize the PR2 Robot\footnote{\url{https://www.willowgarage.com/pages/pr2/overview}} as the robotic platform for our experiments. However, our code is designed to be compatible with multiple robotic platforms and the experiments can be conducted using robots within a real-world setting since it replicates the setup of the real robots in our laboratory as demonstrated by Kazhoyan et al.~\cite{Kazhoyan2020LearningMotion}.

During the interaction, users can interrupt the robot's actions with plan changes, formally defined as minor interruptions (e.g., ``I would like to eat cornflakes instead of bread''), or with major interruptions to stop the system entirely (e.g., ``Stop!'').
The flow of information is managed by a large language model (LLM), acting as a dialogue bridge between the user's natural language input and the robot planner's command execution. Figure~\ref{fig:structure} illustrates our architecture, highlighting the interaction between the speech processing module and the robot planner through this dialogue bridge.

The speech processing module detects voice activity, recognizes speech, and estimates the user's age from the audio stream. The transcribed sentence, age group, and speech confidence level are passed to the dialogue bridge, where the LLM identifies commands and extracts parameters. The user receives confirmation, and the robotic agent performs the requested action according to the user's specifications.
During the action execution, the LLM continuously provides feedback to the user based on the robot's symbolic state, with a frequency determined by the user's age. Communication between the different modules is implemented in ROS, and we publish our code on GitHub.

\subsection{Model}
\begin{figure}[t]
\centering
\includegraphics[width=0.8\textwidth]{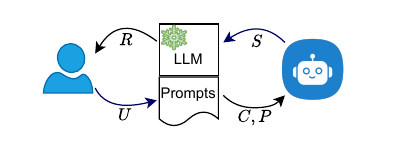}
\caption{The concept of the Dialogue Bridge. The LLM connects the user and the robot by processing the user's utterances (U), turning them into a command (C) with extracted target properties (P) for the robot, as well as monitoring the internal state (S) of the robot and generating an appropriate response (R) to the user.} \label{fig:dialog_bridge}
\end{figure}
\subsubsection{Age and Speech Recognition.}
\label{subsubsec:ageasr}
We utilize Voice Activity Detection (VAD) with Silero VAD\footnote{\url{https://github.com/snakers4/silero-vad}} to eliminate the need for wake-words, allowing for natural interaction with the user. Detected speech segments are processed using the Faster Whisper\footnote{\url{https://github.com/SYSTRAN/faster-whisper}} ASR model (whisper-small, float16 precision) based on Radford et al.~\cite{Radford2022}, alongside an Age Recognition (AR) model. The AR model integrates a pre-trained Whisper Encoder with an attention-based classifier to predict a binary age group (0: young, 1: old) for the dialogue bridge and robot planner. We set the threshold for the binary split at the age group `fifties'.
We detect user traits at every interaction to allow more flexibility for future multi-person scenarios but to prevent oscillations here, we pass along the age averaged over the last five interactions.

The AR model is trained on a modified version of the Common Voice 11.0 \cite{Ardila2019} dataset. We combine the training and validation splits, exclude samples lacking age information (reducing the dataset by about 30\%), and include only speakers with five or more samples. This results in a dataset of 176,448 utterances from 5,217 speakers.
The classification model is trained for 10 epochs with an initial learning rate of 1e-3 and a linear warm-up schedule, using a 70-30 train-test split. 

\subsubsection{Dialogue Bridge.}
\label{dialogue_bridge}
As shown in Figure~\ref{fig:dialog_bridge}, we examine the abilities of LLMs to serve as a dialogue bridge between the user and the robot.
Each turn of the message process can be formally represented:
\begin{equation}
R, C, P = LLM(U, S|prompts)
\end{equation}
where \(U\) denotes the user's utterance and age; 
\(S\) denotes the robot's symbolic states (`step', `interruptable', `move\_arm', `move\_base', `current\_location', and `destination\_location');
\(R\) denotes the response to the user;
\(C\) denotes the commands (minor and major) to the robot, the minor commands include: `bring\_me', `setting\_breakfast', `replace\_object', and `change\_location', the major command is `stop';
\(P\) denotes the target properties of the object (`type', `size', and `color').
While the robot can be interrupted by the user at any time, some of the atomic actions of the robot need to be finished before plan changes can be implemented (e.g. opening the cabinet to look for an object). We use the boolean `interruptable' variable to inform the dialogue bridge of these specific actions. However, major interruptions are the exception and the robot immediately stops the current step, which is why the interaction needs to be restarted after a major interruption occurs.

We prompt the LLM (GPT-3.5) to extract commands and properties from the user's input and provide different levels of feedback depending on the user's age based on the robot's symbolic state. We provide examples of the various commands and their properties. For instance, ``Please bring me a cup instead of a bowl.'' requires the LLM to identify `replace\_object', which is a minor interruption, and `cup' as the replacement object for `bowl'. For older users, the LLM initially generates information about the robot's movement between different locations and arm movements while searching for objects, and the robot is overall more vocal about its actions.
For younger users, the feedback is initially reduced to simple confirmations and the robot displays a higher level of autonomy.

\subsubsection{Robot Planner.}

The robot planner is responsible for the execution of high-level commands passed from the dialogue bridge. The planner is implemented in \mbox{PyCRAM}\footnote{\url{https://github.com/cram2/pycram}}, a framework for developing cognitive robot control programs through symbolic plans. Based on the CRAM cognitive architecture \cite{Beetz2010CRAMCognitive} and adapted for Python3, PyCRAM transforms symbolic plans into concrete parameters guiding robot actions ~\cite{kazhoyan17designators}. This adaptive approach allows the same plan to be used for various tasks, incorporating user feedback without altering the core structure.

The high-level goals provided by the dialogue bridge are handled through designators, which are symbolic descriptions filled at runtime. For example, a designator for picking up an object might specify the object type (e.g., `mug'), the arm to be used, and the grasp type, which consists of a series of atomic actions that represent the high-level action. At execution, the perception module provides the missing details, such as the orientation and placement of an object for grasping.
In the simulation, a placeholder perception module is used alongside the IK solver, which communicates with the same parameters as the real-world equivalents would, to facilitate a switch between simulated and real robots.

Our setup enhances PyCRAM with three key features for immediate responses to dynamic changes or emergencies initiated by a human agent: 1) an Interrupt Client, that allows flexible adjustments (minor interruptions) and shutdown requests (major interruptions), 2) retry and monitor functionalities, which enable recovery actions for plan failures, and 3) dynamic object handling, which allows real-time updates to object states based on the interactions with the user. The robot can navigate to objects, open drawers and doors, and perform pick-and-place actions, all while accommodating plan changes based on user feedback.
These additions to the PyCRAM language enhance the flexibility, robustness, and efficiency of robotic task execution in complex and dynamic environments.

\section{Evaluation and Results}
\begin{figure}[t]
\centering
\includegraphics[width=0.6\textwidth]{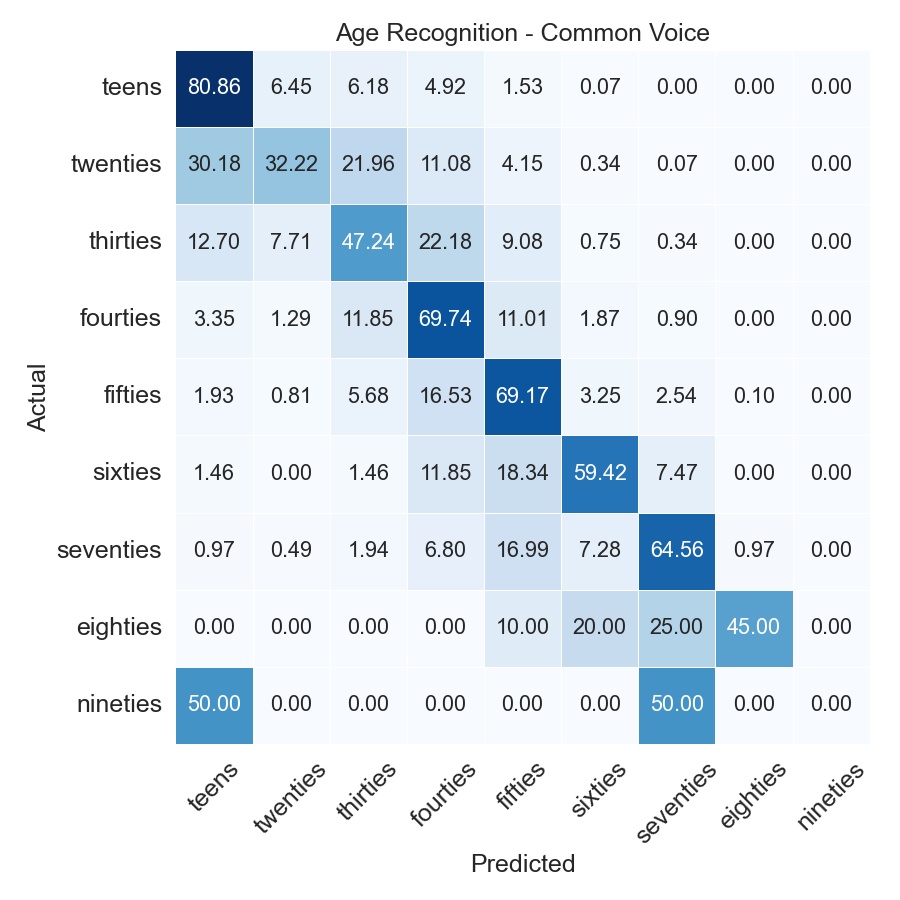}
\caption{The confusion matrix for the Age Recognition model. The matrix shows that the model predicts either the correct age group or one of the two adjacent groups.} \label{fig:cm_ar}
\end{figure}
We evaluate each module in our framework separately and then perform a comprehensive system evaluation using two scenarios with 150 system trials each. We perform the system trials ourselves: 3 users (2 male, 1 female), with age groups `twenties' and `thirties'.
For the first scenario (`bring\_me'), we assess the system's ability to adapt to plan changes with a minor interruption. Initially, the user asks the robot to bring a cup, then interrupts to request a bowl instead. The interaction is considered successful if the robot returns the cup and only the bowl is placed on the table. 
The second scenario (`setting\_breakfast') evaluates the system's response to minor and major interruptions. The user requests breakfast, and while the robot is setting the table, they first ask for a cup as well (minor interruption) and then bring the robot to a standstill by saying ``Stop!'' (major interruption). This scenario is successful if the cup is added to the breakfast items and the system stops as requested.

\subsection{Age and Speech Recognition}
\label{subsec:asrareval}
We measure the performance of the age recognition model by its classification accuracy, as described in Section~\ref{subsubsec:ageasr}. The model can differentiate between older and younger voices with an accuracy of 97.8~\% on the Common Voice dataset. To better illustrate the model's performance, we also include a more detailed evaluation of the nine different age groups (`teens', `twenties', ..., `nineties').
The AR model reaches an accuracy of only 59.5~\% on the Common Voice dataset for the prediction of the nine age groups, but the confusion matrix in Figure~\ref{fig:cm_ar} shows that even in failure cases, the model predicts one of the two adjacent age groups. This behavior is replicated in the system trials, but the AR model classifies the binary age group of the users correctly every time. This will need to be verified in a future user study with older participants and more gender variety.

The speech recognition model is evaluated with word error rate (WER) and character error rate (CER), which measure the number of words or characters transcribed falsely.
The model reaches 14.84~\% WER and 5.20~\% CER on the Common Voice test dataset. 

Since transcripts are not available during the system trials, we instead examine the occurrence of incomplete or erroneous sentences (IES) during the interaction with the user. A sentence is considered incomplete if the transcription is cut off prematurely, and erroneous if the transcription introduces errors to the system (e.g., understanding `pole' instead of `bowl'). We also evaluate the repetition rate (RR), which measures the average number of times the user must repeat a command for it to be executed correctly in one scenario. 

The results show that for `setting\_breakfast' the percentage of IES is on average $25.50\pm\%14.32\%$. For `bring\_me' the mean value is $29.28\%\pm22.95\%$.
The evaluation of the RR shows that for `bring\_me' the user has to repeat themselves on average less than once ($m=0.8644\pm 0.2573$) during successful interactions, which amounts to 75.33\% of all interactions. For `setting\_breakfast' the RR is slightly higher ($m=1.1121\pm 0.9941$), which indicates that on average the user has to repeat themselves more than once in successful interactions (86\%). 

\subsection{Dialogue Bridge}
\begin{table}[t]
    \caption{Command and Object Properties Recognition Average Accuracy(\%){\scriptsize$\pm$Standard Deviation}. `Add object' refers to object requests, and `Delete object' refers to object changes. Type is the object identifier. Color, size, and location are additional properties.}
    \begin{center}
        \scalebox{0.92}{\begin{tabular}{c|c|c|c|c|c|c|c|c|c}
          \hline 
          \multirow{2}{*}{\textbf{LLM}}&\textbf{Command}&\multicolumn{4}{c|}{\textbf{Add object}}&\multicolumn{4}{c}{\textbf{Delete object}}\\
          \cline{3-10}
          &&Type&Color&Size&Location&Type&Color&Size&Location\\
          \hline 
            gpt-3.5-turbo-1106&\textbf{81.57}&\textbf{89.08}&86.60&68.40&84.53&\textbf{83.12}&\textbf{85.54}&\textbf{83.77}&99.96\\              &\scriptsize$\pm$0.003&\scriptsize$\pm$0.004&\scriptsize$\pm$0.003&\scriptsize$\pm$0.001&\scriptsize$\pm$0.001&\scriptsize$\pm$0.003&\scriptsize$\pm$0.001&\scriptsize$\pm$0.002&\scriptsize$\pm$0.000\\
          \hline 
          gpt-3.5-turbo-0125&80.93&82.43&\textbf{88.56}&\textbf{69.28}&\textbf{85.48}&76.48&84.75&82.04&99.99\\              &\scriptsize$\pm$0.003&\scriptsize$\pm$0.002&\scriptsize$\pm$0.003&\scriptsize$\pm$0.003&\scriptsize$\pm$0.004&\scriptsize$\pm$0.004&\scriptsize$\pm$0&\scriptsize$\pm$0.002&\scriptsize$\pm$0.000\\
          \hline 
        \end{tabular}
        }
        \label{tab: dialog bridge evaluation}
    \end{center}
\vspace{-0.6cm}
\end{table}
To quantitatively evaluate the dialogue bridge, we constructed a benchmark dataset comprising five objects (`milk', `bowl', `cereal', `spoon', and `cup') with four colors (`green', `blue', `red', `white'), three sizes (`small', `normal', `big'), and three locations (`countertop', `dishwasher', `cabinet'). 
We collect ten template instructions to request an object (e.g. ``Bring me the small red cup.'') and generate 800 instructions for the command `bring\_me' by combining different object attributes.
Interruptions to replace an object can either be expressed in a single sentence containing all necessary information (e.g. ``Bring me a cup instead of a bowl.''), or require extracting the object from the context of previous instructions (e.g. ``Bring me the bowl instead.''). In this module test, we only consider the first situation. We collect 15 template instructions for replacing one object with another, resulting in 1770 instructions by combining different object attributes. 
Additionally, we collect 41 variants to request breakfast preparation from the robot, leading to 2611 generated instructions for the benchmark dataset overall.

In the module test, we examine two different GPT-3.5 models (gpt-3.5-turbo-1106 and  gpt-3.5-turbo-0125) due to their cost efficiency. Table \ref{tab: dialog bridge evaluation} shows their performance in recognizing command and object properties in the generated instructions. The models perform similarly on the benchmark dataset across three experiments, with overall above-average accuracies and low standard deviations, indicating consistent performance. Because our instructions for replacing objects do not include location details, the accuracy of deleting object locations is nearly 100\%. We decided to use gpt-3.5-turbo-1106 for our system trials, due to its higher performance in command recognition.

During the system trials, the dialogue bridge handles the flow of information between modules, complicating live evaluation. Instead, we document the behavior of the model across the 150 trials. While the system works as intended for a majority of the cases, as discussed in Section~\ref{subsubsec:ageasr}, the most common reasons for unsuccessful interactions are 1) the LLM wrongly classifying the request for object replacement and fetching two objects instead of one, 2) the LLM not responding, and 3) the ASR system sending faulty transcriptions. 

\subsection{Robot Planner}

We evaluate the robot's capability to interrupt and adapt its behavior during transporting tasks by generating permutations of possible commands in the form of ROS messages from a list of objects in our environment (`milk', `bowl', etc.). We ensure that we only evaluate scenarios similar to those encountered in the overall system evaluation to maintain comparable rates of executed actions.
After each command the robot receives, we check its properties and structure and whether it leads to correct robot behavior. During both the individual evaluation and the system trials, we measure the overall performance with the rate of successfully executed commands and the rate of ignored commands. Ignored commands include commands classified as `other', unknown command types, unavailable objects in the environment, or objects not meeting the specified criteria.

The results of the module tests, as shown in Table~\ref{tab: robot eval only}, demonstrate the high success rate of the robot for `bring\_me' and `setting\_breakfast'. In the first scenario, involving fetching and replacing objects, the robot successfully executes 98\% of all received commands. In the second scenario, involving setting the table and stopping the robot, the robot successfully performs 92\% of the received commands. The percentages of ignored commands are due to issues with the IK solver during grasping or the timing of the replace command, such as the command reaching the robot after the fetching action has already been completed. 

For the system trials, the robot receives a larger number of commands overall, due to noise introduced by the repetition rate (RR) and incomplete and erroneous sentence (IES) rate, as well as potential misclassifications by the dialogue bridge. This is reflected in the higher number of ignored commands. 
In the first scenario 88.58\% and in the second scenario 78.47\% of all received communications are either classified as `other' or contain formatting errors and are thereby ignored. However, the overall success rate of the system trials shows the robot executes the plan changes correctly in 75.33\% of all trials for scenario one and in 86\% of all trials for scenario two.

\begin{table}[t]
\centering
\caption{The results of the evaluation of the robot planner. The percentage of ignored commands compared to correctly identified and executed commands is displayed for each scenario. ST Success Rate is the percentage of 150 trials that were executed successfully for each scenario.}
\scalebox{0.92}{\begin{tabular}{c|l|c|c|c}
\hline
\textbf{Scenario} &  \textbf{Command} & \textbf{Robot Evaluation} & \textbf{System Trials (ST)} & \textbf{ST Success Rate}\\
\hline
\multirow{3}{*}{1}  & bring\_me          & 50\% & 7.96\% & \multirow{3}{*}{$75.33\%$}\\ 
                    & replace\_object    & 48\% & 3.46\% &\\ 
                    & ignored    & 2\%  & 88.58\% &\\ \hline
\multirow{4}{*}{2}  & setting\_breakfast & 46\% & 8.10\% & \multirow{4}{*}{$86.00\%$}\\
                    & bring\_me          & 42\% & 9.34\% &\\
                    & stop               & 4\%  & 4.08\% &\\
                    & ignored    & 8\%  & 78.47\%\\ \hline
\end{tabular}}
\label{tab: robot eval only}
\end{table}

\section{Discussion}

In our work, we introduce a framework that uses user traits, such as age, to adapt Human-Robot Interaction (HRI) scenarios to specific user groups and incorporates interruptions to integrate plan changes during action execution. We evaluate our architecture per module and in two scenarios, each with 150 system trials, achieving an overall success rate of 75.33\% for scenario one and 86\% for scenario two.

We chose age as the user-specific trait to modulate interactions within our framework. 
The confusion matrix (Figure~\ref{fig:cm_ar}) and accuracy values on the Common Voice dataset demonstrate that our age recognition (AR) model can reliably identify age ranges rather than specific age groups. The increased confusion in the `nineties' category is mainly due to the limited training data for older age groups, which does not affect our scenario, as we only distinguish between older and younger speakers. The high accuracy in binary age classification (97.8\%) and the AR model's performance in the system trials demonstrate that the model is robust against age-range fluctuations. This should be validated in future user studies since the only age groups presented in the system trials were `twenties' and `thirties'.
During system trials, we observed that the large language model (LLM) was able to distinguish the user's age and generate different responses at the beginning of each interaction. 
However, this behavior diminished after several iterations due to the LLM's memory capacity. As new conversations are appended to the conversation history, the LLM tends to forget the initial prompt.

In addition to providing feedback based on the detected age and robot state, the dialogue bridge is responsible for connecting user input to robot actions. 
The results in Table~\ref{tab: dialog bridge evaluation} demonstrate that the model reliably identifies correct commands and object properties in module-based evaluations.
However, the system trials introduce noise in the form of repetitions and erroneous sentences, leading to a higher number of iterations per interaction. 
The high standard deviation values in the incomplete and erroneous sentence (IES) rate, discussed in Section~\ref{subsubsec:ageasr} indicate that user voice characteristics and microphone quality greatly impact the reliability of voice activity detection. The repetition rate (RR) is influenced not only by the IES rate but also by the rate of command misclassifications by the LLM. In scenario one, repetitions were equally caused by automatic speech recognition (ASR) and LLM errors. In scenario two, the higher RR was more frequently due to transcription errors associated with shorter sentences, which provide less contextual information.

Additionally, in its current state, the LLM passes every user input on to the robot planner, resulting in a higher number of commands being sent overall and an increased rate of ignored commands by the robot.
Since the robot planner performs with high accuracy during the module test (Table~\ref{tab: robot eval only}), we can infer that the lower number of correctly received commands is due to that influx of unclassified utterances. This indicates a need for future iterations of the dialogue bridge to include pre-filtering mechanisms to address faulty or unknown transcriptions from the speech module. Moreover, refining the robot's interrupt handling will further enhance its responsiveness and reliability.

During preliminary trials, we observed that high levels of vocal feedback sometimes interfered with user interruptions when the sound played over the loudspeakers. This issue arose because the speech recognition system is turned off while the robot is speaking to prevent self-talk. For evaluation purposes, we disabled this functionality, but future iterations will address this by implementing approaches to filter out self-talk from the audio stream~\cite{Yue2024}.

\section{Conclusion}
We present a framework for human-robot interaction that leverages interruptions and adaptive feedback to enhance personalization while maintaining a low cognitive load for the user. By using user-specific traits to create intuitive starting points, our system adapts to individual users through feedback and real-time plan adjustments.
Our results demonstrate reliable performance across each module during isolated tests, highlighting the system's modularity. System trials provide an initial exploration into the opportunities and limitations of our architecture, with the framework successfully handling user interruptions and repeated input in 75.33\% of trials for scenario one and 86\% for scenario two. 

Future work will focus on optimizing interruption handling, feedback mechanisms, and recovery procedures to reduce ignored commands and improve the success rate of command classification and message generation. Additionally, extending experiments to real-world settings through user studies with diverse age groups will further validate our simulation results and ensure the robustness of the framework in diverse environments.

We believe that our framework will significantly contribute to the development of more intuitive and user-friendly HRI systems, ultimately enhancing their practical application in everyday scenarios.

\begin{credits}
\subsubsection{\ackname} The authors gratefully acknowledge funding from Horizon Europe under the MSCA grant agreement No 101072488 (TRAIL), the China Scholarship Council (CSC), and the German Research Foundation DFG under project CML (TRR 169), LeCAREbot, and as part of the Collaborative Research Center (Sonderforschungsbereich) 1320 Project-ID 329551904 (EASE). The authors would also like to thank Matthias Kerzel for his proofreading and constructive suggestions on this paper.

\subsubsection{\discintname} The authors have no competing interests to declare that are relevant to the content of this article.

\end{credits}
%
%
%
\bibliographystyle{splncs04}
\bibliography{mybibfile}
\end{document}